\documentclass[letterpaper]{article} 
\usepackage[T1]{fontenc}
\usepackage{aaai2026}  
\usepackage{times}  
\usepackage{helvet}  
\usepackage{courier}  
\usepackage[hyphens]{url}  
\usepackage{graphicx} 
\usepackage{natbib}  
\usepackage{caption} 
\usepackage{algorithm}
\usepackage{algorithmic}
\usepackage{enumitem}
\usepackage{multirow}
\usepackage{amsmath}
\usepackage{newfloat}
\usepackage{listings}
\DeclareCaptionStyle{ruled}{labelfont=normalfont,labelsep=colon,strut=off} 
\floatstyle{ruled}
\newfloat{listing}{tb}{lst}{}
\floatname{listing}{Listing}
\title{Lost in Benchmarks? Rethinking Large Language Model Benchmarking with Item Response Theory}
\def\@fnsymbol#1{\ensuremath{\ifcase#1\or *\or \dagger\or \ddagger\or
   \mathsection\or \mathparagraph\or \|\or **\or \dagger\dagger
   \or \ddagger\ddagger \else\@ctrerr\fi}}
\newcommand{\ssymbol}[1]{^{\@fnsymbol{#1}}}
\author{
    Hongli Zhou\textsuperscript{\rm 1}\equalcontrib,
    Hui Huang\textsuperscript{\rm 1}\equalcontrib, 
    Ziqing Zhao\textsuperscript{\rm 1}, 
    Lvyuan Han\textsuperscript{\rm 1}, 
    Huicheng Wang\textsuperscript{\rm 1}, \\
    \textbf{Kehai Chen\textsuperscript{\rm 2}, }
    \textbf{Muyun Yang\textsuperscript{\rm 1}$\thanks{Corresponding authors.}$, }
    \textbf{Wei Bao\textsuperscript{\rm 3}$\ssymbol{2}$, }
    \textbf{Jian Dong\textsuperscript{\rm 3}$\ssymbol{2}$, }
    \textbf{Bing Xu\textsuperscript{\rm 1}, } \\
    \textbf{Conghui Zhu\textsuperscript{\rm 1}, }
    \textbf{Hailong Cao\textsuperscript{\rm 1}, }
    \textbf{Tiejun Zhao\textsuperscript{\rm 1}}
}
\affiliations{

    \textsuperscript{\rm 1}Faculty of Computing, Harbin Institute of Technology, Harbin, China \\
    \textsuperscript{\rm 2}School of Computer Science and Technology, Harbin Institute of Technology, Shenzhen, China \\
    \textsuperscript{\rm 3}China Electronics Standardization Institute, Beijing, China \\
    hongli.joe@stu.hit.edu.cn, yangmuyun@hit.edu.cn, \{baowei, dongjian\}@cesi.cn

}

\usepackage{bibentry}

\begin{document}

\maketitle

\begin{abstract}
The evaluation of large language models (LLMs) via benchmarks is widespread, yet inconsistencies between different leaderboards and poor separability among top models raise concerns about their ability to accurately reflect authentic model capabilities. This paper provides a critical analysis of benchmark effectiveness, examining mainstream prominent LLM benchmarks using results from diverse models. We first propose Pseudo-Siamese Network for Item Response Theory (PSN-IRT), an enhanced Item Response Theory framework that incorporates a rich set of item parameters within an IRT-grounded architecture. PSN-IRT can be utilized for accurate and reliable estimations of item characteristics and model abilities. Based on PSN-IRT, we conduct extensive analysis on 11 LLM benchmarks comprising 41,871 items, revealing significant and varied shortcomings in their measurement quality. Furthermore, we demonstrate that leveraging PSN-IRT is able to construct smaller benchmarks while maintaining stronger alignment with human preference.
\end{abstract}

\begin{links}
    \link{Code}{https://github.com/Joe-Hall-Lee/PSN-IRT}
\end{links}

\section{Introduction}
As the scale and performance of large language models (LLMs) continue to grow, accurately measuring their capabilities has become increasingly important \cite{10.1145/3641289, hongli-etal-2024-mitigating, huang-etal-2025-empirical}. Currently, the performance of LLMs is primarily evaluated through various benchmarks \cite{NEURIPS2024_ad236edc}, which are comprehensive test suites consisting of carefully designed questions to assess model behavior across different tasks. However, in practice, existing benchmarks often exhibit significant limitations, prompting consideration of their effectiveness \cite{mcintosh2024inadequacies}.

\begin{figure}[t]
    \centering
        \includegraphics[width=1.00\linewidth]{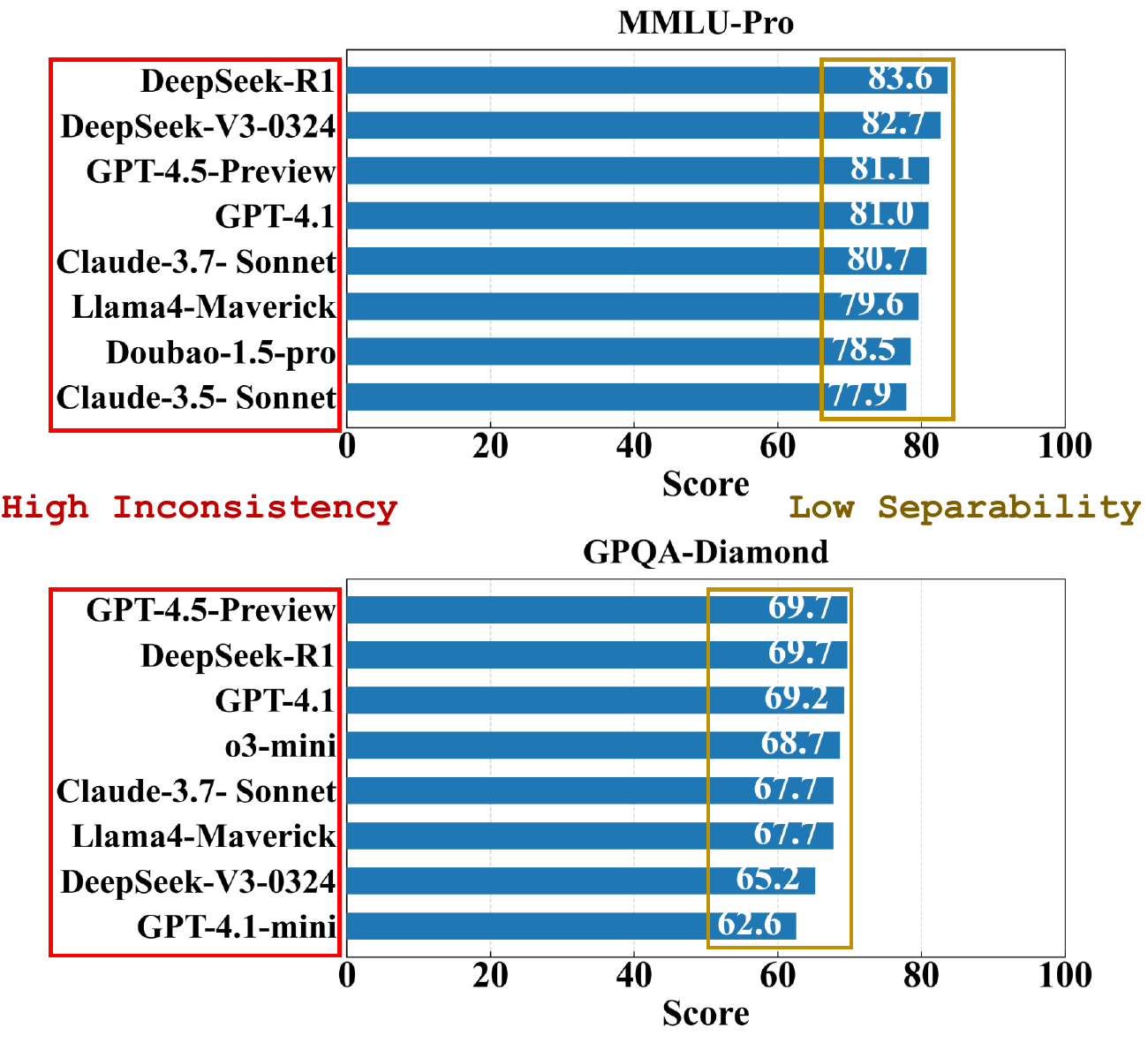}
        \caption{Illustration of weak separability and ranking inconsistencies in LLM benchmarks\protect\footnotemark.}
    \label{fig:intro-fig}
\end{figure}

As illustrated in Figure \ref{fig:intro-fig}, on one hand, even benchmarks that are designed to measure similar underlying capabilities often produce inconsistent leaderboards, leading to substantial ranking variations for the same model \cite{perlitz2024llmbenchmarksagreefixing}. On the other hand, many benchmarks show weak separability among top models, limiting the ability to understand performance differences and further model refinement. Given these limitations, there is a growing need for a more systematic analysis of LLM benchmarks.

\footnotesize\footnotetext{\url{https://opencompass.org.cn}}

In this paper, we conduct a comprehensive analysis for various popular LLM benchmark datasets. Our experiment is based on Item Response Theory (IRT) \cite{lord1968statistical, baker2004item}, a psychometric framework widely used in educational assessment to analyze item properties such as difficulty by modeling item response against examinee ability\footnote{In the paper, an item refers to a benchmark sample.}. However, traditional IRT struggles with the complexity and scale of modern datasets, and its assumption of the normal distribution of abilities does not always hold. 

To advance our objective,  we propose the Pseudo-Siamese Network for IRT (PSN-IRT), a framework for comprehensive and interpretable analysis of benchmark test sets. PSN-IRT processes model identifiers and item identifiers through independent neural network pathways. These pathways respectively estimate latent model ability and a rich set of item parameters. The estimated properties are then integrated using an IRT-based formulation to predict outcome probabilities. This architectural design ensures both powerful modeling capabilities and strong theoretical interpretability.

To validate the effectiveness of PSN-IRT, we conduct extensive experiments on evaluation results from 12 LLMs across 11 benchmarks. Our results show that PSN-IRT outperforms previous approaches in both parameter estimation accuracy and reliability. Based on PSN-IRT, we perform in-depth analyses of these benchmarks using key metrics, with main findings as follows:

\begin{enumerate}[leftmargin=*, itemsep=1mm, parsep=0pt]
    \item LLM benchmarks fail to achieve simultaneous excellence across multiple measurements.
    \item LLM benchmarks suffer from widespread saturation and insufficient difficulty ceilings, limiting their ability to challenge and accurately evaluate advanced models.
    \item LLM benchmarks exhibit data contamination in numerous items, compromising their reliability.
\end{enumerate}

Furthermore, we find that model rankings derived from high-quality datasets selected by PSN-IRT are more consistent with human preference and offer stronger discriminability among top models.

Our contributions are summarized as follows:
\begin{enumerate}[leftmargin=*, itemsep=1mm, parsep=0pt]
\item We propose PSN-IRT, a benchmark analysis framework with superior estimation accuracy and reliability.
\item We use PSN-IRT for an in-depth analysis of mainstream benchmarks and find that current LLM benchmarks present deficiencies in many aspects.
\item We show that selecting items with PSN-IRT leads to model rankings that better align with human preference.
\end{enumerate}

\begin{figure*}[!t]
    \centering
        \includegraphics[width=0.98\linewidth]{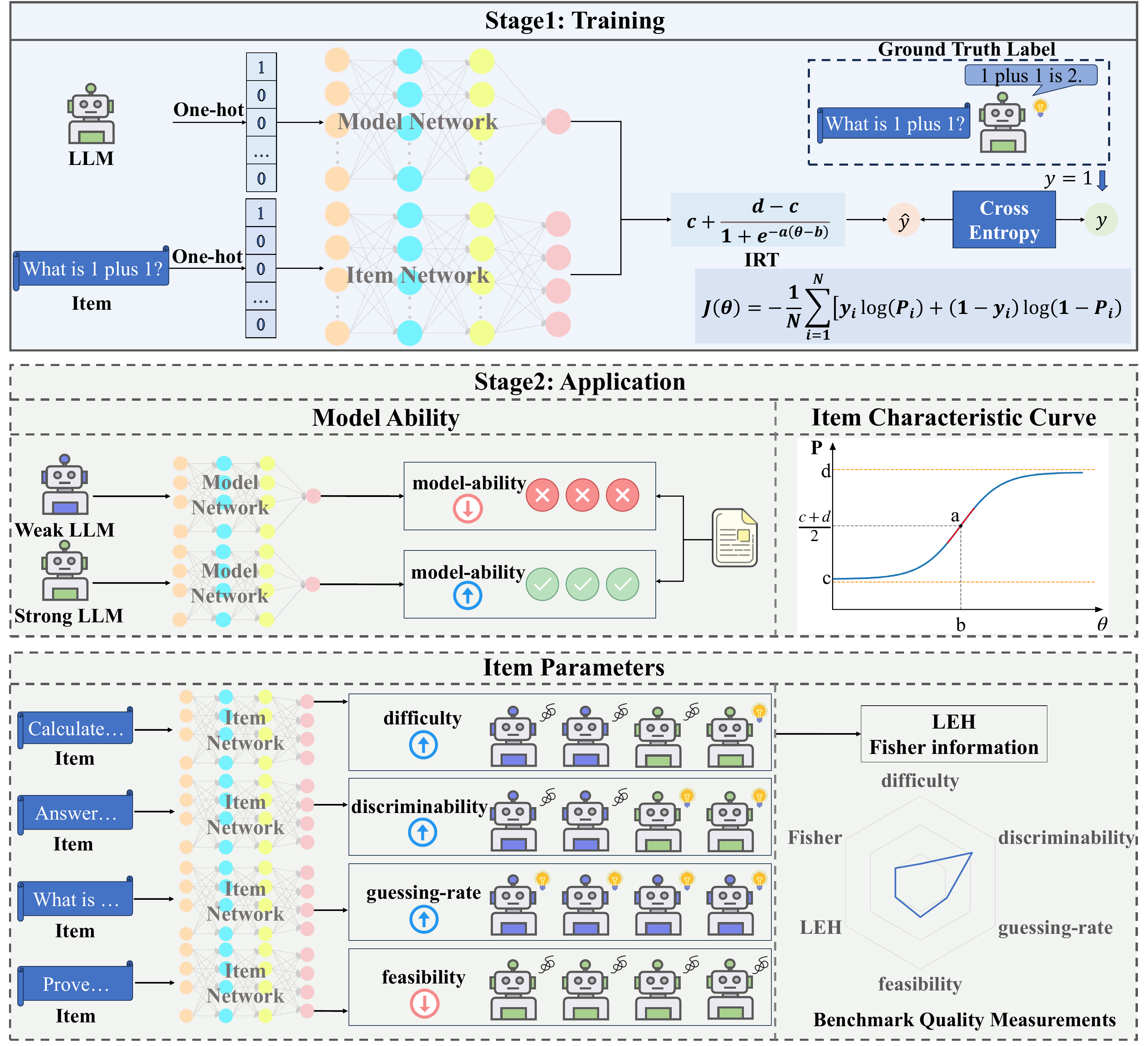}
        \caption{The illustration of our proposed PSN-IRT. Separate neural networks estimate model-ability ($\theta$) and item parameters ($a,b,c,d$), which are then combined via the IRT formula to predict the probability of a correct response. After that, the networks can be leveraged for estimating properties for models or items, respectively.}
    \label{figure:main-fig}
\end{figure*}

\section{Background}

\subsection{Related Work}

Researchers have explored various methods to evaluate and analyze the quality of LLM benchmarks. Recent efforts have introduced quantitative metrics that operate at a holistic, dataset level. For example, Benchmark Agreement Testing (BAT) assesses benchmark reliability by comparing the consistency of model rankings across different leaderboards \cite{perlitz2024llmbenchmarksagreefixing, white2025livebench}. Other novel metrics have also been proposed for more nuanced dataset-level analysis \cite{li2024crowdsourced}. Delving deeper than macro-level comparisons, another significant line of work focuses on content-based analysis to ground evaluation in concrete capabilities. For instance, ADeLe \cite{zhou2025general} leverages scalable cognitive rubrics and item demand features to build highly interpretable capability profiles for AI systems.

Item Response Theory (IRT) \cite{lord1968statistical, baker2004item} offers a psychometric framework for data-driven, item-level analysis. Within natural language processing (NLP), IRT has been primarily utilized to diagnose fundamental item properties such as difficulty and discriminability \cite{vania-etal-2021-comparing, byrd-srivastava-2022-predicting}. More recently, its principles have also been used to inspire methods for improving evaluation efficiency, for instance, through optimized item selection \cite{pmlr-v235-maia-polo24a} or adaptive testing paradigms \cite{pmlr-v267-zhuang25e, pmlr-v267-truong25c}. However, due to the complexity and scale of modern datasets and its assumption of normally distributed abilities, the application of IRT on LLM benchmarks is still underexplored \cite{ye2025large}.

\subsection{Preliminary: Item Response Theory}

Item Response Theory (IRT) \cite{lord1968statistical} is a psychometric framework widely used in educational and cognitive assessments to model the relationship between benchmark items and examinee abilities. Unlike Classical Test Theory (CTT)~\cite{devellis2006classical}, which relies on aggregate test scores, IRT analyzes individual item responses as a function of a latent ability $\theta$, enabling precise measurement of both item and examinee properties.

Central to IRT is the Item Characteristic Curve (ICC), a mathematical function that describes the probability of a correct response, $P(X=1 \mid \theta)$, as a function of ability $\theta$. Typically logistic, the ICC visualizes how item properties influence performance, serving as the foundation for IRT.

The simplest IRT model, the One-Parameter Logistic (1PL) model, defines the ICC as:

\begin{equation}
P(X=1 \mid \theta) = \frac{1}{1 + e^{-(\theta - b)}}
\end{equation}

\noindent
where $b$ denotes item difficulty. When $\theta = b$, the probability of the examinee generating a correct response is 0.5. The 1PL assumes all items have equal discriminability, a simplification that may not hold in complex testing scenarios.

More advanced IRT models introduce additional parameters to capture diverse item properties. However, traditional IRT often suffer from inaccurate parameter estimation and rely on assumptions like normal ability distributions, which may not align with real-world data \cite{tsutsumi2021deep}.

\section{Pseudo-Siamese Network for IRT}

In this section, we propose the Pseudo-Siamese Network for IRT (PSN-IRT), designed to diagnose benchmark item quality by analyzing LLM responses. Specifically, one property can be inferred for each model: \textit{model-ability}, and four properties can be inferred for each benchmark item: \textit{discriminability}, \textit{difficulty}, \textit{guessing-rate} and \textit{feasibility}.

\paragraph{Model Architecture.} The PSN-IRT architecture comprises two independent networks: a model network and an item network. The model network processes one-hot encoded LLM identifiers to estimate \textit{model-ability} $\theta$, while the item network handles one-hot encoded item identifiers to produce the four IRT parameters: \textit{discriminability} $a$, \textit{difficulty} $b$, \textit{guessing-rate} $c$, and \textit{feasibility} $d$. 

Each network consists of three fully connected layers with ReLU activation functions, enabling efficient and stable parameter estimation. After that, these estimated properties are then processed by a Logistic Calculation Layer. This layer operates using the Four-Parameter Logistic (4PL) model, an advanced IRT formulation that extends simpler models like the 1PL to capture nuanced behaviors:

\begin{equation}
P(X=1 \mid \theta) = c + (d - c) \cdot \frac{1}{1 + e^{-a(\theta - b)}}
\label{equation:P}
\end{equation}

\noindent
where each parameter controls a different aspect of benchmark item behavior:

\begin{itemize}[leftmargin=*, itemsep=1mm, parsep=0pt]
    \item The difficulty $b$ determines the ability level where the probability changes most significantly.
    \item The discriminability $a$ reflects an item's power to differentiate between models of varying abilities. 
    \item The guessing-rate $c$ captures how likely models are to succeed without full understanding.
    \item The feasibility $d$ represents the maximum probability that even highly proficient models will correctly answer the item.
\end{itemize}

Notably, PSN-IRT follows standard IRT assumptions such as unidimensionality and monotonicity, which generally hold for LLM benchmarking~\cite{kipnis2025metabench, pmlr-v267-truong25c}.

\paragraph{Training and Applications.} The PSN-IRT is trained end-to-end using the observed binary response data in the form of (Model, Item, Response, Outcome), where the binary outcome indicates the correctness of each LLM on each benchmark item. During training, PSN-IRT processes encoded pairs of LLM and item identifiers through their respective network pathways. For each pair, the PSN-IRT framework first internally estimates the specific model and item properties. These estimated properties are then passed to a Logistic Calculation Layer, with the training objective as follows:
\begin{small}
    \begin{equation}
        J(\theta) = - \frac{1}{N} \sum_{i=1}^{N} \left[ y_i \log(P_i) + (1 - y_i) \log(1 - P_i) \right]
    \end{equation}
\end{small}

\noindent
where $N$ is the batch size. The training objective is to minimize the discrepancy between the model's prediction and the observed binary outcomes. Consequently, all learnable weights within both the model and item networks are updated simultaneously, concurrently optimizing the estimated properties for both models and benchmark items.

Upon completion of training, the two networks of PSN-IRT can be used for evaluating model performance and diagnosing benchmark characteristics, respectively. Specifically, the model network, by processing a model's encoding, outputs an estimate of the model-ability $\theta$. Meanwhile, the item network, given a benchmark item's encoding, outputs its four estimated psychometric parameters $a, b, c, d$, which can be used for an in-depth diagnosis of individual item characteristics and the overall benchmark quality.
\section{Efficacy of PSN-IRT}

\subsection{Experimental Setup}
\label{sec:experimental-setup}
\paragraph{Datasets.} To evaluate existing LLM benchmarks, we select 11 datasets that vary in domain coverage. They serve as a basis for comparing different IRT-based methods and analyzing test set properties. An overview of the datasets is provided in Table~\ref{tab:datasets}.

To construct training data, we conduct evaluations using OpenCompass \cite{2023opencompass}. The evaluation results are collected in the form of binary outcomes, indicating whether each model answered each item correctly. Then, we construct a unified outcome matrix by aggregating item-level results across all models and benchmarks. We split the interactions into training, validation, and test sets.

\begin{table*}[!tb]
\centering
\begin{tabular}{lccccc}
\hline
\textbf{Benchmark} & \textbf{Data Size} & \textbf{Domain} & \textbf{Metric} & \textbf{Format} \\ \hline
ARC-C \cite{allenai:arc}                 & 295            & General           & EM          & Multiple Choice \\
BBH \cite{suzgun-etal-2023-challenging}     & 6,511          & General           & EM          & Mixed \\
Chinese SimpleQA \cite{he2024chinesesimpleqachinesefactuality} & 3,000  & Knowledge         & LLM-as-a-Judge & QA \\
GPQA Diamond \cite{rein2024gpqa}         & 198            & Science           & EM          & Multiple Choice \\
GSM8K \cite{cobbe2021gsm8k}              & 1,319          & Math              & EM          & QA \\
HellaSwag \cite{zellers-etal-2019-hellaswag} & 10,042       & General           & EM          & Multiple Choice \\
HumanEval \cite{chen2021codex}           & 164            & Code              & Pass@1      & QA \\
MATH \cite{hendrycksmath2021}            & 5,000          & Math              & EM          & QA \\
MBPP \cite{austin2021programsynthesislargelanguage} & 500     & Code              & Pass@1      & QA \\
MMLU \cite{hendryckstest2021}            & 14,042         & General           & EM          & Multiple Choice \\
TheoremQA \cite{chen-etal-2023-theoremqa} & 800            & Science           & EM          & QA \\ \hline
\end{tabular}
\caption{Benchmarks used in this paper.}
\label{tab:datasets}
\end{table*}

\begin{table*}[t]
\centering
\begin{tabular}{ccccccc}
\hline
\textbf{Model} & \textbf{Parameter} & \textbf{Method} & \textbf{ACC} & \textbf{F1} & \textbf{AUC} & \textbf{Kendall's $\tau$} \\ \hline

\multirow{4}{*}{IRT} & \multirow{4}{*}{4PL} & MLE    & 0.7211 & 0.8034 & 0.7012 & 0.9697 \\
                     &                    & MCMC   & 0.7070 & 0.7811 & 0.7278 & 0.9697 \\
                     &                    & VI     & 0.7201 & 0.8015 & 0.6940 & 0.9091 \\
                     &                    & VIBO   & 0.7188 & 0.8007 & 0.7055 & 0.9697 \\ \cline{1-7}

Deep-IRT & 1PL & Deep Learning & 0.7974 & 0.8516 & \textbf{0.8519} & 0.9697 \\
PSN-IRT  & 4PL & Deep Learning & \textbf{0.7998} & \textbf{0.8538} & 0.8485 & \textbf{1.0000} \\ \hline

\end{tabular}
\caption{Comparison of prediction accuracy and rank reliability across different methods.}
\label{tab:results}
\end{table*}

We mainly focus on currently high-performing models, including 360GPT2-Pro\footnote{\url{https://ai.360.cn/}}, DeepSeek-V3\footnote{\url{https://platform.deepseek.com/}}, Doubao-pro\footnote{\url{https://www.volcengine.com/product/doubao}}, Gemini-1.5\footnote{\url{https://ai.google.dev/}}, Hunyuan-Turbo\footnote{\url{https://hunyuan.tencent.com/}}, Moonshot-v1\footnote{\url{https://platform.moonshot.cn/}}, Qwen-Plus\footnote{\url{https://qwen.ai/apiplatform}}, and Yi-Lightning\footnote{\url{https://platform.lingyiwanwu.com/}}. However, testing only strong models might obscure differences in the discriminability of the test sets, as most items could be solved easily, thus limiting a comprehensive evaluation of test set quality \cite{martinez2019item}. To address this, we intentionally included several relatively weaker models, including Gemma-2B-it \cite{gemmateam2024gemmaopenmodelsbased}, Mistral-7B-Instruct-v0.1 \cite{jiang2023mistral7b}, Qwen2.5-3B-Instruct \cite{qwen2025qwen25technicalreport}, and Vicuna-7B-v1.3 \cite{vicuna2023}.

\paragraph{Baselines.} We mainly compare PSN-IRT with traditional IRT methods, namely IRT estimated based on traditional parameter estimation techniques, including Maximum Likelihood Estimation (MLE), Markov Chain Monte Carlo (MCMC) \cite{hastings1970monte}, Variational Inference (VI) \cite{jordan1999introduction}, and VIBO \cite{WuDDPG20}. We also compare with Deep-IRT \cite{tsutsumi2021deep}, which is an extension of 1PL IRT that learns item-trait interactions via deep networks.

\paragraph{Metrics.} To assess whether the learned parameters meaningfully reflect corresponding properties, we use two quantitative metrics:

\begin{itemize}[leftmargin=*, itemsep=1mm, parsep=0pt]
\item \textbf{Prediction accuracy}: Following standard practice in educational testing \cite{eignor2013standards}, we assess how well a method predicts whether a model answers an item correctly, reporting accuracy, F1 score, and ROC AUC.
\item \textbf{Rank stability}:  To evaluate the reliability of model ranking induced by each method, we split the test set into two subsets, estimate model abilities separately, and compute Kendall’s~$\tau$ between the resulting rankings.
\end{itemize}

\subsection{Main Results}

Table \ref{tab:results} shows the estimation accuracy and reliability of PSN-IRT compared to traditional IRT and Deep-IRT. Both Deep-IRT and PSN-IRT significantly outperform traditional IRT in prediction accuracy. Despite its simpler Logistic output layer rooted in the IRT formula, PSN-IRT attains prediction accuracy on par with the more complex Deep-IRT, and surpasses it in rank reliability.

Given our goal to comprehensively capture diverse item characteristics, we adopt the more expressive 4PL structure for all methods except Deep-IRT. Experiments also confirm that PSN-IRT performs best with 4PL.

\subsection{Ablation Study}
\label{sec:ablation}
To assess whether more complex architectures offer benefits over our straightforward design, we conduct an ablation study. Specifically, we compare PSN-IRT with two variants: one replaces one-hot inputs with semantic embeddings from Instructor \cite{su-etal-2023-one}, and the other uses a GNN \cite{SU2022109547} instead of the MLP backbone.

As shown in Table~\ref{tab:ablation}, both alternatives underperform our original design. We argue that semantic similarity between item texts does not necessarily reflect similarity in measurement characteristics. For the GNN variant, its neighborhood aggregation mechanism risks diluting the unique statistical signals of individual models and items, thereby hindering precise parameter estimation. These results suggest that in this estimation scenario, preserving the individuality of models and items is more effective than introducing architectural complexity.

\begin{table}[!t]
\centering
\begin{tabular}{lcccc}
\hline
\textbf{Method} & \textbf{ACC} & \textbf{F1} & \textbf{AUC} & \textbf{Kendall} \\
\hline
PSN-IRT    &  \textbf{0.7998} & \textbf{0.8538} & \textbf{0.8485} & \textbf{1.0000} \\
\quad + Embedding  & 0.7808 & 0.8413 & 0.8310 & 0.9394 \\
\quad + GNN         & 0.7928 & 0.8490 & 0.8197 & 0.7273 \\
\hline
\end{tabular}
\caption{Ablation study on PSN-IRT's input representation and network architecture.}
\label{tab:ablation}
\end{table}

Furthermore, since PSN-IRT estimates parameters that are specific to the training data, requiring retraining to analyze new items, we also investigate the stability of these estimates when the item pool changes. To simulate this, we conduct an experiment by independently training separate PSN-IRT models on random subsets of the items. We then compute the Pearson correlation between the item parameters estimated from these subsets and those estimated from the full dataset.

As shown in Table \ref{tab:stability_results}, the results reveal a high degree of correlation across all parameters. This indicates that PSN-IRT learns highly consistent intrinsic properties for the items regardless of the specific data sample, demonstrating the stability of parameter estimation.

\begin{table}[t]
\centering
\begin{tabular}{lcccc}
\hline
\textbf{Data} & \textbf{Difficulty} & \textbf{Discrim.} & \textbf{Guessing} & \textbf{Feasibility} \\
\hline

30\% & 0.9009 & 0.8186 & 0.8274 & 0.9025 \\
50\% & 0.7437 & 0.7835 & 0.8776 & 0.9330 \\
70\% & 0.9519 & 0.7414 & 0.8320 & 0.9442 \\
\hline
\end{tabular}
\caption{Pearson correlation coefficients of item parameters estimated on random subsets of the data versus those estimated on the full dataset. All correlations are statistically significant ($p < 0.0001$).}
\label{tab:stability_results}
\end{table}

\begin{table*}[!tb]
\centering
\begin{tabular}{lccccccc}
\hline
\textbf{Dataset} & \textbf{Difficulty} & \textbf{Discriminability} & \textbf{Guessing} & \textbf{Feasibility} & \textbf{LEH} & \textbf{Fisher} & \textbf{Total} \\ \hline
Chinese SimpleQA & 2  & 3  & 1  & 9  & 3  & 5  & 23 \\
TheoremQA        & 1  & 10 & 3  & 11 & 2  & 2  & 29 \\
MBPP             & 5  & 6  & 4  & 8  & 4  & 4  & 31 \\
MATH             & 4  & 1  & 2  & 7  & 7  & 10 & 31 \\
GPQA Diamond     & 3  & 11 & 6  & 10 & 1  & 1  & 32 \\
BBH              & 6  & 5  & 8  & 6  & 6  & 7  & 38 \\
MMLU             & 9  & 9  & 9  & 5  & 5  & 3  & 40 \\
GSM8K            & 8  & 2  & 5  & 3  & 11 & 11 & 40 \\
HumanEval        & 7  & 4  & 7  & 4  & 9  & 9  & 40 \\
HellaSwag        & 10 & 7  & 10 & 2  & 8  & 6  & 43 \\
ARC-C            & 11 & 8  & 11 & 1  & 10 & 8  & 49 \\ \hline
\end{tabular}

\caption{Ranks of datasets based on average item-level properties estimated by PSN-IRT. Each rank is obtained by averaging the property over all items in a dataset and then sorting these averages; smaller rank values correspond to better performance.}
\label{tab:dataset_ranks}
\end{table*}

\begin{figure*}[t]
    \centering
    \includegraphics[width=1.0\linewidth]{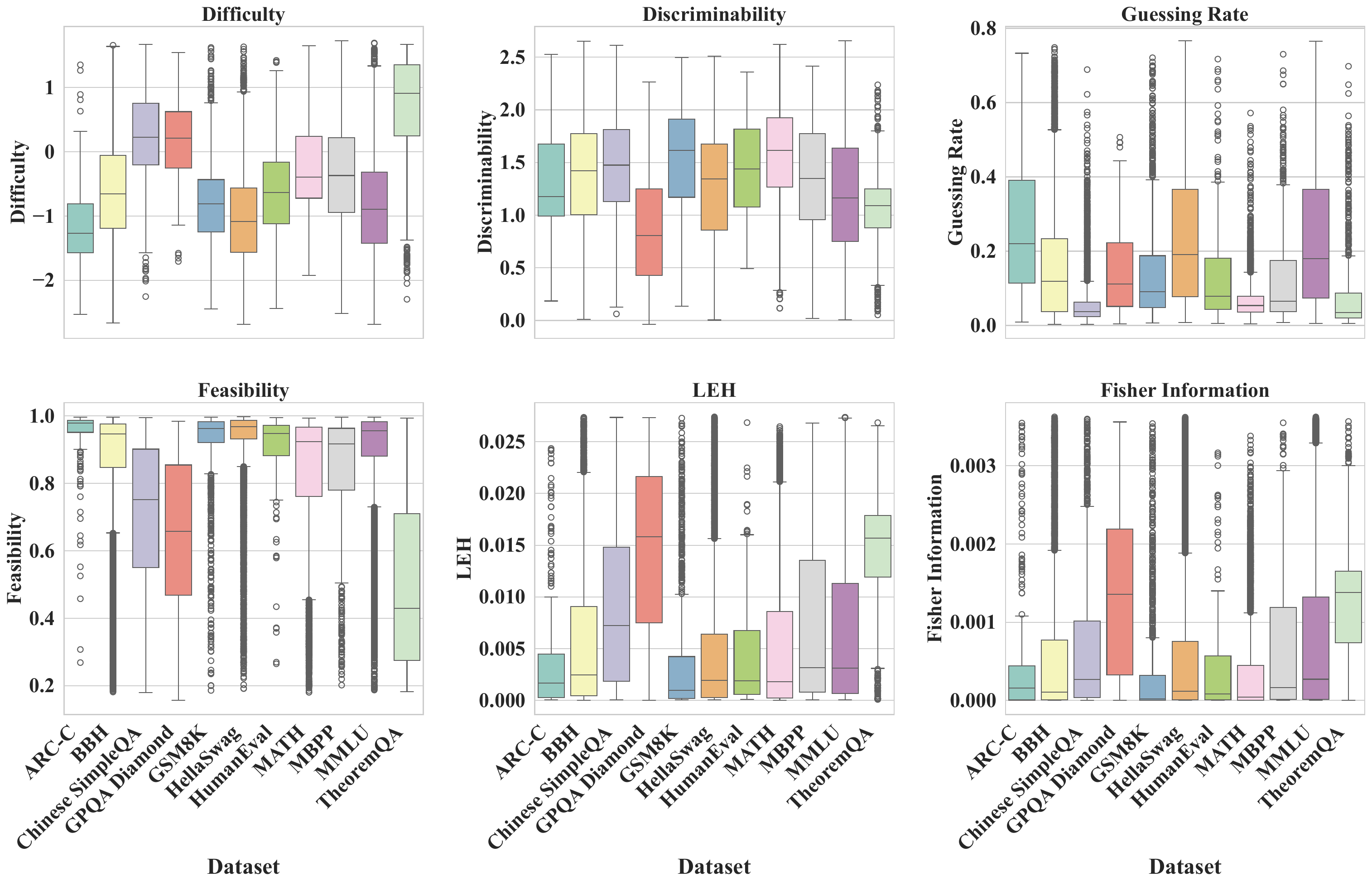}
    \caption{Distribution of item-level properties across 11 LLM benchmarks.}
    \label{fig:benchmark_analysis}
\end{figure*}

\section{Benchmark Analysis with PSN-IRT}
\label{sec:benchmark-analysis}

\subsection{Benchmark Quality Measurements}
In this section, we delve into a detailed analysis of the benchmarks based on different measurements. Specifically, we leverage the 4 item parameters (difficulty, discriminability, guessing-rate and feasibility) estimated by PSN-IRT, along with two additional metrics for evaluating item quality: Local Efficiency Headroom (LEH) \cite{vania-etal-2021-comparing} and Fisher information \cite{lord1980applications}. 

LEH score assesses the potential of a test example to evaluate future progress in LLMs. It is calculated as the derivative of the item's Item Characteristic Curve (ICC) with respect to the highest observed latent ability. A high LEH score indicates that even the top model remains far from the saturation point of the ICC.

Fisher information measures the amount of information an item provides about a model's ability level. Items with higher Fisher information are more informative for estimating model abilities. For 4PL IRT, it is defined as:

\begin{equation}
I(\theta) = \frac{a^2 (P(\theta) - c)^2 (d - P(\theta))^2}{(d - c)^2 P(\theta)(1 - P(\theta))},
\end{equation}

\noindent
where $P(\theta)$ is defined as Equation \ref{equation:P}.

\subsection{Benchmark Analysis}

Based on the measurements, we provide an aggregate rank for each benchmark, calculated from the sum of individual ranks in Table \ref{tab:dataset_ranks}, where a lower total rank value indicates better overall quality. Additionally, Figure \ref{fig:benchmark_analysis} visualizes the distribution of item-level properties across the 11 LLM benchmarks. Building on these experimental results, our detailed analysis is presented as follows.

\paragraph{Finding 1: LLM benchmarks fail to achieve simultaneous excellence across multiple measurement properties.} As shown in Table \ref{tab:dataset_ranks}, the results clearly demonstrate this lack of simultaneous excellence. While Chinese SimpleQA achieves the best overall score, no benchmark performs universally well across the individual metrics. For instance, Chinese SimpleQA itself suffers from poor feasibility with a rank of 9. This pattern of strengths being offset by significant weaknesses is common across benchmarks, which underscores the inherent challenges and trade-offs in benchmark design. Moreover, the property distributions visualized in Figure \ref{fig:benchmark_analysis} further highlight this variability and the difficulty for any single benchmark to achieve balanced, high quality across all desired measurement properties.

\paragraph{Finding 2: Current LLM benchmarks exhibit an insufficient difficulty ceiling to challenge the most advanced models.} As shown in Table \ref{tab:dataset_ranks}, difficulty is varied among different benchmarks. Datasets such as TheoremQA, Chinese SimpleQA, and GPQA Diamond lead in difficulty and continue to challenge many LLMs. In contrast, other benchmarks including ARC-C, HellaSwag, and MMLU have become comparatively easy, where many of the items are readily solved by high-performing models, diminishing their utility for differentiating top-tier systems.

Crucially, as shown in Figure~\ref{fig:benchmark_analysis}, even the most challenging items in existing benchmarks exhibit relatively low difficulty levels. For instance, on the IRT scale, the highest item difficulty values rarely exceed 1.0, while top models such as DeepSeek-V3 have estimated abilities well above 3.0. This indicates that even the hardest questions are significantly below the capability level of elite LLMs, highlighting a lack of sufficient challenge for frontier capabilities.

\paragraph{Finding 3: Generally low LEH scores across most benchmarks reveal widespread item saturation.} While Table~\ref{tab:dataset_ranks} indicates that datasets like GPQA Diamond and TheoremQA achieve the highest relative ranks for average LEH, suggesting they offer more headroom than other benchmarks, the absolute LEH values in Figure \ref{fig:benchmark_analysis} are often smaller than ideal. Conversely, benchmarks such as GSM8K and ARC-C exhibit lower average LEH scores. This signifies that current high-performing models are already approaching or have reached the performance ceiling for a substantial portion of items within these test sets. This necessitates the development of new benchmark items and datasets explicitly designed with greater headroom as models continue to evolve.

\paragraph{Finding 4: Outlier-high guessing-rates in many benchmark items flag risks of data contamination.} The guessing-rate reflects the chance that a model can answer a question correctly without actual understanding, typically by exploiting format-based cues or shortcuts. As shown in Figure \ref{fig:benchmark_analysis}, datasets like ARC-C, HellaSwag, and MMLU generally present higher guessing-rates, possibly due to their multiple-choice formats. Notably, further inspection of the item guessing-rate parameter distributions, as depicted in Figure \ref{fig:benchmark_analysis}, reveals that a considerable number of items across several benchmarks exhibit unusually high guessing-rates. This observation regards the potential for such items to indicate data contamination, namely the content or answers for these items might have been present in the models' pre-training data \cite{pmlr-v267-zhuang25e}.

\paragraph{Finding 5: Widespread low-feasibility items in scientific benchmarks reveal flawed question design.} Feasibility estimates how well a question can be answered based on the information provided.  Low feasibility often arises from underspecified or overly broad questions, where even a strong model cannot determine a clear answer. As shown in Figure~\ref{fig:benchmark_analysis}, datasets like TheoremQA and GPQA Diamond exhibit lower feasibility, likely due to complex scientific contexts or vague problem descriptions. In contrast, ARC-C and HellaSwag have high feasibility, suggesting their questions are well-formed and specific. Low feasibility more often reflects weaknesses in annotation quality or task design, and such items can distort evaluation by penalizing models for failing to resolve ambiguities \cite{rodriguez-etal-2021-evaluation}.

\section{Efficient Benchmarking with PSN-IRT}

While the pursuit of comprehensive LLM evaluation has led to increasingly large benchmarks, sheer item volume neither guarantees quality nor comes without significant computational cost. This section, therefore, explores how PSN-IRT can facilitate more efficient benchmarking. We investigate whether strategically curating smaller, highly informative item sets can yield model comparisons that maintain or even surpass the accuracy and reliability of larger, undifferentiated collections.

\paragraph{Experimental Setup.} To evaluate the effectiveness of different item selection strategies, we design an experiment to evaluate the effectiveness of different item selection strategies. Our methodology involves Benchmark Agreement Testing (BAT) \cite{perlitz2024llmbenchmarksagreefixing}. We first establish a reference model ranking by aggregating results from two public leaderboards: Chatbot Arena \cite{pmlr-v235-chiang24b} and OpenCompass Arena \cite{2023opencompass}\footnote{These arenas are based on human evaluation and thus reflect human preference regarding overall model capabilities.}.

Subsequently, various subsets of benchmark items are chosen based on different criteria. Model rankings generated using these subsets are then compared against the reference arena ranking using Kendall's $\tau$ to measure agreement. Alongside agreement, we also consider the variance of model scores on these subsets; higher variance generally indicates the subset's capacity to differentiate more clearly between models. The evaluations are conducted using two distinct model groupings: one set comprising a mix of stronger and weaker models ("w/ weak models"), and another set from which weaker models are excluded ("w/o weak models"), to specifically assess separability among stronger contenders, with weak models introduced in Section \ref{sec:experimental-setup}.

\paragraph{Results.} As shown in Table \ref{tab:selection}, selecting items based on Fisher information consistently produces model rankings with superior alignment to the reference arena ranking. For instance, a carefully selected subset composed of just 1,000 items chosen via the Fisher information criterion achieves a Kendall's $\tau$ of up to 0.9048. This level of agreement markedly surpasses that achieved by rankings derived from using all available items or from similarly sized randomly selected subsets.

\begin{table}[t]
\setlength{\tabcolsep}{1.5mm}
\centering
\begin{tabular}{lcccc}
\hline
\multirow{2}{*}{\textbf{Method}} & \multicolumn{2}{c}{\textbf{w/ weak models}} & \multicolumn{2}{c}{\textbf{w/o weak models}} \\
\cline{2-5}
& \textbf{Variance} & \textbf{Kendall} & \textbf{Variance} & \textbf{Kendall} \\
\hline
All & 0.0434 & 0.6444 & 0.0010 & 0.2381 \\
\hline
\multicolumn{5}{l}{\textit{Top 400}} \\
\hline
Random & 0.0421 & 0.6444 & 0.0007 & 0.2381 \\
Clustering & 0.0308 & 0.6000 & \textbf{0.0058} & 0.1429 \\
Discrim. & \textbf{0.1841} & 0.5556 & 0.0000 & 0.2381 \\
Fisher & 0.0049 & \textbf{0.6889} & 0.0033 & \textbf{0.7143} \\
\hline
\multicolumn{5}{l}{\textit{Top 1000}} \\
\hline
Random & 0.0406 & 0.6444 & 0.0009 & 0.2381 \\
Clustering & 0.0166 & 0.6444 & \textbf{0.0041} & 0.2381 \\
Discrim. & \textbf{0.1805} & 0.6000 & 0.0000 & 0.3813 \\
Fisher & 0.0039 & \textbf{0.6889} & 0.0030 & \textbf{0.9048} \\
\hline
\end{tabular}
\caption{Variance and Kendall's $\tau$ for various data selection strategies with and without weak models.}
\label{tab:selection}
\end{table}

In contrast, other item selection approaches are less effective for distinguishing strong models. For instance, selecting items based on discriminability yields zero score variance when weak models are excluded, indicating it merely separates broad capability tiers rather than providing granularity among top-performers. Separately, clustering items based on their success/failure vectors \cite{pacchiardi2024100instancesneedpredicting} also fails to improve ranking correlation, even when it increases score separability.

\section{Conclusion}
In this work, we introduce PSN‑IRT, a framework that combines psychometrics with neural networks to provide comprehensive and reliable analyses of LLM benchmarks. Through extensive experiments on 12 LLMs across 11 diverse datasets, we demonstrate that current benchmarks often suffer from uneven measurement properties, insufficient difficulty ceilings, item saturation, and data contamination. Moreover, we show that strategically selecting smaller, high‑quality item sets using PSN‑IRT enhances alignment with human preference and separability among top models.

\section*{Acknowledgments}
This work is supported by National Natural Science Foundation of China (62276077, 62376075, 62376076), Department of Science and Technology of Heilongjiang (Grant No. ZY04JD04), and the Key Laboratory of Computing Power Network and Information Security, Ministry of Education (Grant No. 2023ZD027).

\bibliography{aaai2026}

\setcounter{secnumdepth}{2}
\renewcommand{\thesection}{\Alph{section}}
\renewcommand{\thesubsection}{\thesection.\arabic{subsection}}
\appendix
\clearpage
\onecolumn

\section{Implementation Details}
\subsection{Estimation Ability of PSN-IRT}
To evaluate the estimation ability of PSN-IRT against baselines, we first construct a unified response matrix by aggregating item-level results across all models and benchmarks. We then split these model-item interactions into training (60\%), validation (20\%), and test sets (20\%). The model is trained on the training set using the Adam optimizer with a learning rate of 0.003 and a weight decay of 0.0001, and a batch size of 512. To ensure robust model selection and prevent overfitting, we employ an early stopping strategy with a patience of 5 epochs, monitored based on the F1 score on the validation set.

For the subsequent benchmark analysis presented in Section \ref{sec:benchmark-analysis}, where the goal is to obtain the most stable parameter estimates for all items, we train the PSN-IRT model on the entire unified dataset. In this phase, the model is trained for a fixed 30 epochs using the same hyperparameters. All training is conducted on a single NVIDIA A800-80G GPU.
\subsection{Ablation Study}
\label{sec:appendix-ablation}
This section provides additional implementation details for the ablation study variants mentioned in Section \ref{sec:ablation}.

\paragraph{Semantic Embedding Variant.}
To explore the utility of item content, we substitute one-hot encoded item identifiers with dense semantic embeddings. These embeddings are generated for the text of each benchmark item using the pre-trained Instructor-XL\footnote{https://huggingface.co/hkunlp/instructor-xl}. They then serve as the direct input to the item network, replacing the one-hot encoding pathway while the rest of the PSN-IRT architecture remained unchanged.

However, these variants fail to yield significant improvements. We will explore improved input representations and model architectures as future work to enhance model robustness.

\paragraph{Graph Neural Network (GNN) Variant.}
To investigate the role of network architecture, we implemented a custom GNN-based encoder. This encoder updates a model's initial embedding by aggregating information from all interactions in the training graph. For each interaction between a model $s$ and an item $q$ with a binary outcome $rel \in \{0, 1\}$, a cognitive degree $\mathbf{l}_{s,q}$ is first computed:
\begin{equation}
\mathbf{l}_{s,q} = \text{Linear}_{\text{gate}}([\mathbf{e}_q, \mathbf{W}_{rel} \cdot rel])
\end{equation}
where $\mathbf{e}_q$ is the item's initial embedding, $\mathbf{W}_{rel}$ is a trainable weight matrix that projects the scalar response $rel$ into the embedding space, and $[\cdot, \cdot]$ denotes concatenation. This combined vector is then processed by a single linear layer ($\text{Linear}_{\text{gate}}$) to form the cognitive degree. Subsequently, a cognitive state $\mathbf{c}_{s,q}$ is formed through an element-wise product:
\begin{equation}
\mathbf{c}_{s,q} = \mathbf{e}_q \odot \mathbf{l}_{s,q}
\end{equation}
The updated embedding for model $s$, denoted $\mathbf{e}'_s$, is obtained by performing an attention-weighted aggregation of the cognitive states from all items $\mathcal{N}_s$ it has interacted with:
\begin{equation}
\mathbf{e}'_s = \mathbf{e}_s + \sum_{q_i \in \mathcal{N}_s} \alpha_{s,q_i} \cdot \mathbf{c}_{s,q_i}
\end{equation}
where the attention weights $\alpha_{s,q_i}$ are calculated via a softmax over scores. These scores are derived by applying a linear layer and a LeakyReLU activation to a concatenation of the model's embedding $\mathbf{e}_s$ and each cognitive state $\mathbf{c}_{s,q_i}$. A symmetric process is used to update item embeddings.

However, both variants fail to yield performance gains. We leave the exploration of more specialized architectures and feature representations as future work.

\section{Additional Analysis}
\subsection{Performance Comparison across IRT Parmaters}
\label{sec:appendix-additional}

To investigate the impact of model configurations, we evaluate PSN-IRT and baseline methods across 1PL to 4PL parameters, as shown in Table \ref{tab:detail-results}. Among these models, the Two-Parameter Logistic (2PL) model builds upon the 1PL by incorporating an item discriminability parameter $a$:

\begin{equation}
P(X=1 \mid \theta, a, b) = \frac{1}{1 + e^{-a(\theta - b)}}
\end{equation}

Subsequently, the Three-Parameter Logistic (3PL) model extends the 2PL by adding a guessing-rate parameter $c$:
\begin{equation}
P(X=1 \mid \theta, a, b, c) = c + (1 - c) \cdot \frac{1}{1 + e^{-a(\theta - b)}}
\end{equation}

The results demonstrate that performance exhibits no consistent pattern with increasing parameter complexity. While PSN-IRT with 4PL achieves the highest performance, simpler models occasionally outperform more complex ones in traditional IRT methods.

\begin{table*}[!tb]
\centering
\begin{tabular}{ccccccc}
\hline
\textbf{Model} & \textbf{Method} & \textbf{Parameter} & \textbf{ACC} & \textbf{F1} & \textbf{AUC} & \textbf{Kendall's $\tau$} \\ \hline
\multirow{16}{*}{IRT} & \multirow{4}{*}{MLE} & 1PL & 0.7219 & 0.8035 & 0.6970 & 1.0000 \\
                      &                      & 2PL & 0.7172 & 0.7998 & 0.6963 & 0.9091 \\
                      &                      & 3PL & 0.7218 & 0.8031 & 0.7057 & 0.9697 \\
                      &                      & 4PL & 0.7211 & 0.8034 & 0.7012 & 0.9697 \\ \cline{2-7}
                      & \multirow{4}{*}{MCMC} & 1PL & 0.7168 & 0.8011 & 0.7268 & 0.9697 \\
                      &                       & 2PL & 0.7232 & 0.8130 & 0.7301 & 0.9697 \\
                      &                       & 3PL & 0.7172 & 0.8081 & 0.7297 & 1.0000 \\
                      &                       & 4PL & 0.7070 & 0.7811 & 0.7278 & 0.9697 \\ \cline{2-7}
                      & \multirow{4}{*}{VI}   & 1PL & 0.7209 & 0.8024 & 0.6822 & 0.9697 \\
                      &                       & 2PL & 0.7198 & 0.8012 & 0.6938 & 1.0000 \\
                      &                       & 3PL & 0.7189 & 0.8011 & 0.6956 & 0.9697 \\
                      &                       & 4PL & 0.7201 & 0.8015 & 0.6940 & 0.9091 \\ \cline{2-7}
                      & \multirow{4}{*}{VIBO} & 1PL & 0.7007 & 0.7801 & 0.6901 & 0.9697 \\
                      &                       & 2PL & 0.7301 & 0.8048 & 0.6987 & 0.8788 \\
                      &                       & 3PL & 0.7172 & 0.7993 & 0.7048 & 0.9091 \\
                      &                       & 4PL & 0.7188 & 0.8007 & 0.7055 & 0.9697 \\ \hline
\multirow{1}{*}{Deep-IRT} & Deep Learning & 1PL & 0.7974 & 0.8516 & \textbf{0.8519} & 0.9697 \\ \hline
\multirow{4}{*}{PSN-IRT} & \multirow{4}{*}{Deep Learning} & 1PL & 0.7948 & 0.8497 & 0.8473 & 0.9091 \\
                         &                                & 2PL & 0.7560 & 0.8266 & 0.8259 & 0.9697 \\
                         &                                & 3PL & 0.7965 & 0.8510 & \textbf{0.8510} & 0.8788 \\
                         &                                & 4PL & \textbf{0.7998} & \textbf{0.8538} & 0.8485 & \textbf{1.0000} \\ \hline
\end{tabular}
\caption{Comparison of prediction accuracy and rank reliability across different methods.}
\label{tab:detail-results}
\end{table*}

\subsection{Detailed Benchmark Statistics}
\label{sec:appendix-statistics}
We provide the full list of average item-level property estimates for each dataset in Table~\ref{tab:item_property_statistics}. These values are computed as the mean of all item parameters within a benchmark and include difficulty, discriminability, guessing-rate, feasibility, LEH score, and Fisher information. The table allows for fine-grained comparisons between benchmarks beyond their aggregated rankings.

\begin{table*}[h]
\centering
\begin{tabular}{lcccccc}
\hline
\textbf{Dataset} & \textbf{Difficulty} & \textbf{Discriminability} & \textbf{Guessing} & \textbf{Feasibility} & \textbf{LEH} & \textbf{Fisher} \\
\hline
ARC-C             & -1.1637 & 1.2822 & 0.2621 & 0.9481 & 0.0041 & 0.0005 \\
BBH               & -0.6033 & 1.3845 & 0.1657 & 0.8747 & 0.0057 & 0.0006 \\
Chinese SimpleQA  &  0.2656 & 1.4474 & 0.0609 & 0.7088 & 0.0089 & 0.0006 \\
GPQA Diamond      &  0.1902 & 0.8598 & 0.1518 & 0.6437 & 0.0142 & 0.0014 \\
GSM8K              & -0.7899 & 1.5255 & 0.1491 & 0.9167 & 0.0038 & 0.0004 \\
HellaSwag          & -1.0353 & 1.2920 & 0.2449 & 0.9304 & 0.0047 & 0.0006 \\
HumanEval          & -0.6351 & 1.4459 & 0.1565 & 0.8957 & 0.0046 & 0.0005 \\
MATH               & -0.2396 & 1.5793 & 0.0673 & 0.8290 & 0.0054 & 0.0004 \\
MBPP               & -0.3456 & 1.3625 & 0.1345 & 0.8221 & 0.0071 & 0.0007 \\
MMLU               & -0.8383 & 1.1968 & 0.2381 & 0.8921 & 0.0067 & 0.0008 \\
TheoremQA          &  0.6548 & 1.0656 & 0.0888 & 0.5012 & 0.0145 & 0.0013 \\
\hline
\end{tabular}
\caption{Average item-level properties across datasets. Each value represents the mean of the corresponding metric over all benchmark items.}
\label{tab:item_property_statistics}
\end{table*}

\subsection{Relationship Between Item Difficulty and Discriminability}
\label{sec:appendix-diff-disc-analysis}

\begin{figure*}[t]
    \centering
    \includegraphics[width=\textwidth]{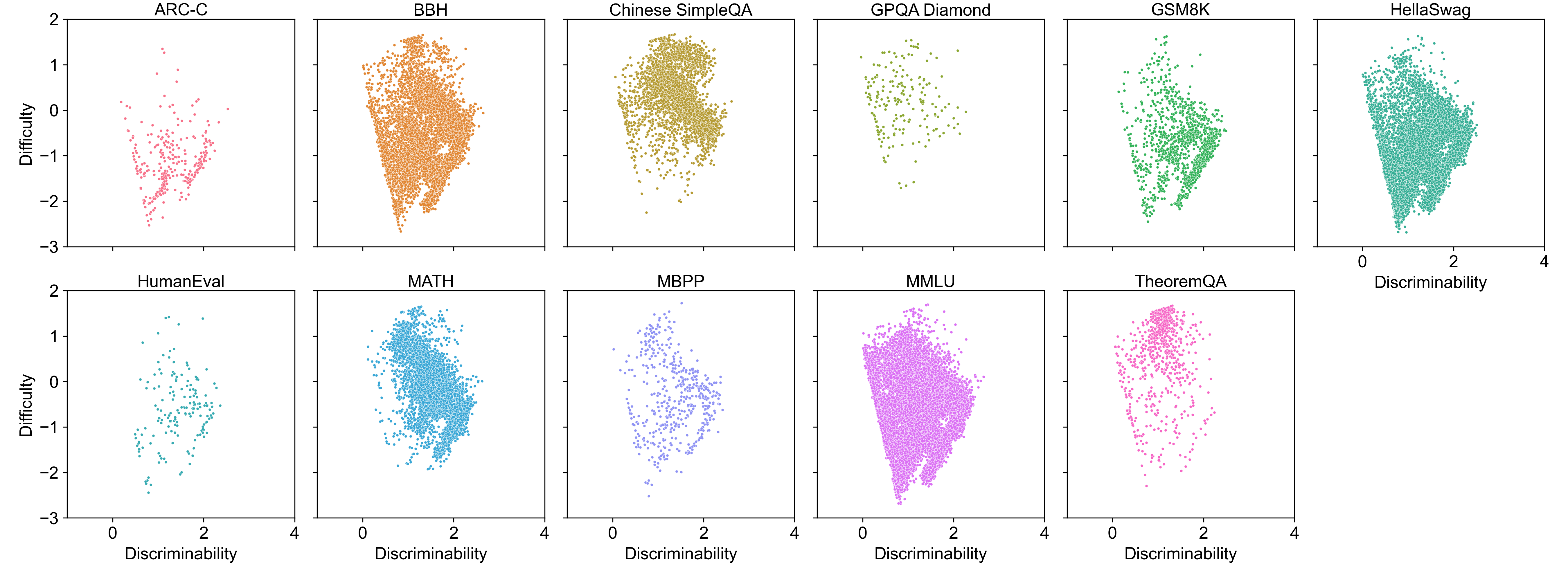}
    \caption{Scatter plots showing the relationship between item difficulty and discriminability across 11 benchmarks. Each plot highlights how difficulty affects discriminability within a dataset.}
    \label{fig:diff_disc}
\end{figure*}

Our analysis reveals a notable relationship between item difficulty and discriminability. As illustrated by the scatter plots in Figure \ref{fig:diff_disc}, item discriminability tends to be lower when item difficulty reaches extreme levels—that is, when items are either very easy or very hard. Consequently, datasets that feature a more balanced distribution of difficulty often exhibit higher overall discriminability, enabling them to more effectively distinguish between models of varying capabilities.

This pattern is also reflected in the dataset-level averages presented in Table \ref{tab:dataset_ranks} in the main text. For instance, benchmarks such as MATH and GSM8K, which contain a range of difficulties, achieve high average discriminability. Conversely, datasets like GPQA Diamond and TheoremQA, which are characterized by a concentration of very difficult items, exhibit the lowest average discriminability. This suggests that their capacity for precise model differentiation is reduced, in part, due to this difficulty-discriminability trade-off. These observations underscore the importance of balancing difficulty when designing test items to ensure a benchmark is not only challenging but also highly discriminative.

\subsection{Case Study}

\begin{figure*}[t]
    \centering
    \includegraphics[width=\textwidth]{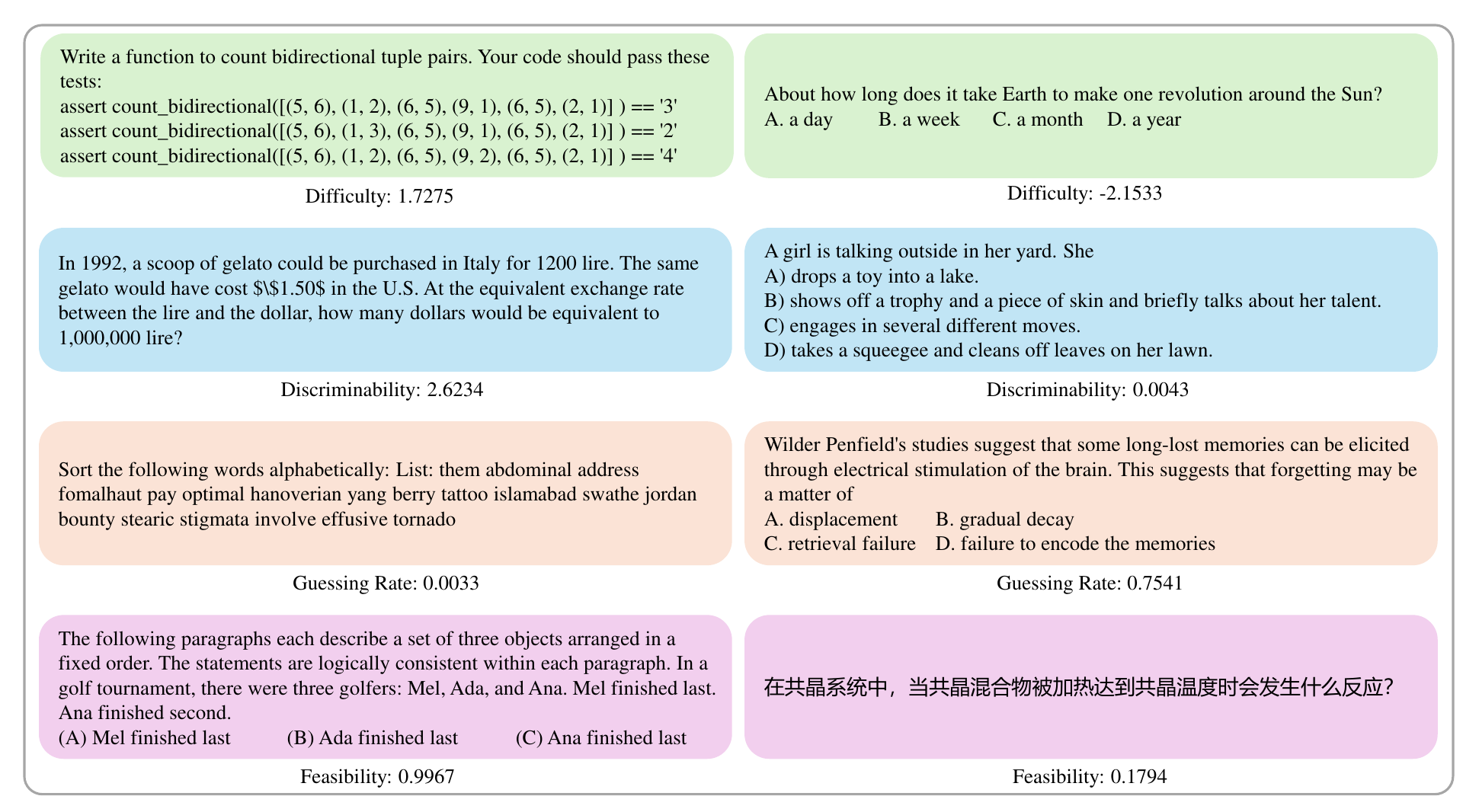}
    \caption{Examples of benchmark items exhibiting high and low estimated values for key psychometric parameters, as determined by PSN-IRT. Each pair shows an item with a high parameter value alongside one with a low value for the specified characteristic.}
    \vspace{-3mm}
    \label{fig:case}
\end{figure*}

As depicted in Figure \ref{fig:case}, these parameters capture a wide range of item behaviors:

\begin{itemize}
\item  For difficulty, the figure contrasts a complex coding task with a simple factual recall question, showcasing items at opposite ends of the challenge spectrum.
\item  For discriminability, the examples include a quantitative reasoning problem with a high value, indicating strong differentiating power, and a common-sense inference task with a low value, suggesting limited ability to distinguish between models.
\item For guessing-rate, an alphabetical sorting task exhibits a very low estimated parameter, offering little opportunity for chance success. In contrast, a multiple-choice question concerning scientific studies shows an exceptionally high guessing parameter. This item's content is publicly accessible online, suggesting an associated risk of data contamination.
\item   For feasibility, a logical arrangement task shows high feasibility, implying clarity and high attainability for proficient models, while a domain-specific question exhibits low feasibility, suggesting potential item ambiguities.
\end{itemize}

\end{document}